\documentclass{article} % For LaTeX2e
\usepackage{preprint,times}

\usepackage{amsmath,amsfonts,bm}

\def\eqref#1{equation~\ref{#1}}
\def\1{\bm{1}}

\DeclareMathAlphabet{\mathsfit}{\encodingdefault}{\sfdefault}{m}{sl}
\SetMathAlphabet{\mathsfit}{bold}{\encodingdefault}{\sfdefault}{bx}{n}

\usepackage{hyperref}
\usepackage{url}
\usepackage{booktabs}
\usepackage{graphicx}
\usepackage{float}
\usepackage{tabularx}
\usepackage{wrapfig}
\usepackage{caption}
\usepackage[table]{xcolor}
  \usepackage{multirow}
  \usepackage{pgfplotstable}
  \pgfplotsset{compat=1.18}
\usepackage{pifont}
\usepackage{xcolor}
\usepackage{bbm}
\definecolor{cYes}{rgb}{0.30,0.76,0.52}
\newcommand{\yes}{\textcolor{cYes}{\ding{51}}}
\newcommand{\no}{\textcolor{black!25}{\ding{55}}}
\usepackage{listings}
\usepackage{booktabs}
\usepackage{tabularx}
\usepackage[breakable,skins,listings]{tcolorbox}

\definecolor{promptbg}{RGB}{248,248,248}
\definecolor{promptborder}{RGB}{220,220,220}
\newtcblisting{promptbox}[1]{%
  enhanced,
  breakable,
  listing only,
  colback=promptbg,
  colframe=promptborder,
  colbacktitle=promptbg,
  coltitle=black,
  fonttitle=\small\sffamily\bfseries,
  title={#1},
  title after break={#1\enspace\normalfont\small (continued)},
  boxrule=0.4pt,
  arc=2pt,
  left=9pt, right=9pt, top=7pt, bottom=7pt,
  before skip=10pt, after skip=10pt,
  listing options={%
    language={},
    basicstyle=\footnotesize\ttfamily,
    columns=fullflexible,
    keepspaces=true,
    breaklines=true,
    breakindent=1em,
    prebreak={}, postbreak={},
    showspaces=false,
    showstringspaces=false,
    showtabs=false,
    tabsize=2,
    literate={—}{{\textemdash}}1
  }
}

\title{EviRover: Reinforcing Agentic Perception Beyond a Glance}

\author{%
  \textbf{Kaixuan Fan$^{1}$ \quad Kaituo Feng$^{1}$ \quad Tianshuo Peng$^{1}$ \quad Yilei Jiang$^{1}$}\\
  \textbf{Manyuan Zhang$^{1}$ \quad Junke Wang$^{2}$ \quad Xiangyu Yue$^{1,\dagger}$}\\[4pt]
  \textnormal{$^{1}$MMLab, The Chinese University of Hong Kong}\\
  \textnormal{$^{2}$Fudan University}\\[2pt]
  \textnormal{$^{\dagger}$Corresponding author}\\[4pt]
  \textnormal{\url{https://github.com/kxfan2002/evirover}}%
}

\definecolor{cGroup}{gray}{0.90}

\definecolor{cLoc}{rgb}{1.00,0.70,0.58}   % peach     -> Localization
\definecolor{cRec}{rgb}{0.62,0.90,0.72}   % mint      -> Recognition
\definecolor{cSpot}{rgb}{0.76,0.65,0.96}  % lavender  -> Spot Diff
\definecolor{cSeg}{rgb}{0.60,0.84,0.97}   % sky blue  -> Segmentation
\definecolor{cCnt}{rgb}{0.98,0.92,0.48}   % butter    -> Counting

\begin{document}
\maketitle

\begin{abstract}

Visual perception is conventionally formulated as a one-shot prediction from a single glance at the image, under the assumption that the image content and the model's parametric knowledge suffice to resolve the query.
This assumption often fails in real-world scenarios that hinge on fine-grained visual details or require knowledge-intensive and up-to-date information.
We term such cases \textit{perception under insufficient evidence} and formulate perception as an agentic process that can obtain information beyond a single glance.
To address the absence of data for this setting, we design two dedicated data generation pipelines, yielding EviRover-SFT-5K and EviRover-RL-12K for training.
We further construct EviLens, a human-verified benchmark comprising 688 instances across five perception categories.
Building on these data, we present EviRover, to our knowledge the first perception agent explicitly trained to resolve perceptual queries through interaction, using supervised fine-tuning followed by agentic reinforcement learning.
Experiments show that the 4B EviRover outperforms its backbone by 30 points on average on EviLens, reaching performance comparable to advanced proprietary models.
The gains transfer beyond EviLens to WebEyes, conventional perception benchmarks, and general multimodal benchmarks, including a 15-point improvement on BrowseComp-VL.
All code, models, and data are released.
\end{abstract}
\section{Introduction}

% spot diff 可以加进来吗？

Multimodal large language models (MLLMs) have progressed rapidly in reasoning and tool use~\citep{ma2025one,feng2026onethinker,du2026towards,yan2026prommsearchagent}. 
These advances have enabled multimodal agents to address increasingly complex visual questions through multi-step interaction~\citep{huang2026vision,fan2026sophiavl,chu2026redsearcher,jiao2026searcheyes}. 
Yet in most such systems, agentic interaction serves question answering rather than perceptual prediction.
Perception tasks themselves have largely retained their conventional formulation~\citep{wang2026locateanything,tang2026visual,wang2026x,deng2026geosam2,pacini2026countingdino}.
Grounding, segmentation, and counting are still commonly treated as one-shot predictions from an image-query pair, based on a single \textit{glance} at the input image.
This formulation assumes that the image and the model's parametric knowledge are sufficient to resolve the query.
In practice, fine-grained visual details may require closer inspection, while identifying the target may depend on knowledge beyond the image.
We refer to such cases as \textit{perception under insufficient evidence}.
In these settings, a single \textit{glance} is insufficient to determine the required perceptual output, yet the model must still produce a location, a boundary, or a count without the opportunity for further search.

We therefore reformulate perception under insufficient evidence as an evidence-seeking process. 
Rather than predicting directly from the initial observation, the model can search through finer-grained visual inspection or external knowledge sources to resolve the perceptual task.
Whereas existing multimodal agents~\citep{huang2026vision,wu-etal-2026-mmsearch,zhang2025thyme} gather evidence to support a textual answer, the proposed formulation treats the perceptual output itself as the objective.
This makes perception a stricter test of evidence seeking: in question answering, searched evidence can be converted directly into a textual answer, whereas in perception it must be translated into a location, a boundary, or a count on the input image, which demands fine-grained visual understanding.
Moreover, the perceptual output is directly verifiable against spatial or numerical annotations, and is correct only if the agent both acquires the missing evidence and correctly relates it to the visual content.
Perception is also of considerable practical value, as accurate perception underpins downstream applications such as embodied manipulation~\citep{zhang2024sam,Kim_2026_CVPR} and image editing~\citep{Liu_2024_CVPR,Liu_2026_CVPR}.
Despite these advantages, agentic perception poses a central challenge: the missing evidence varies across queries. A small target calls for closer inspection, whereas an unfamiliar identity calls for external search. Deciding what to acquire and how is therefore part of the task rather than a fixed procedure.
Recent prompt-based workflows~\citep{yang2026web,tang2026rose} have begun to incorporate web search into perception.
However, they rely on manually designed prompting strategies at inference time, so their searching behavior is bounded by human-designed heuristics rather than learned from task feedback.
\citet{liang2026segresearch} train an agent that interleaves reasoning with web search. But it is limited to segmentation and assumes that missing evidence is always external knowledge.

Inspired by recent developments in agentic reinforcement learning for visual question answering~\citep{zheng2026deepeyes,wu-etal-2026-mmsearch,hong2026deepeyesv2}, we ask: \textit{can we train agents to learn when and how to search to push perception beyond a single glance?}

\begin{figure}[t]
    \centering
    \includegraphics[width=\linewidth]{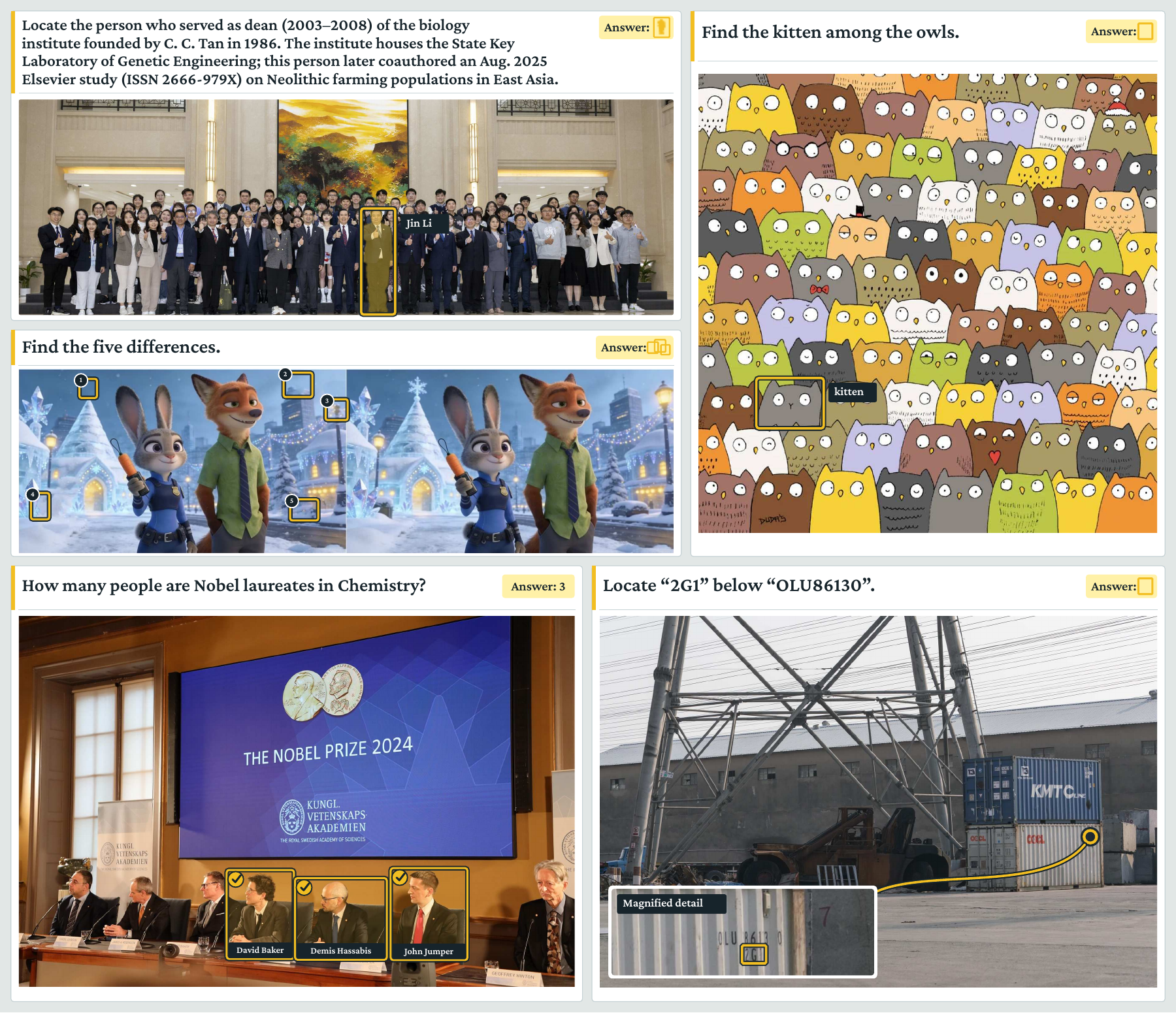}
    \vspace{-10pt}
    \caption{Representative examples from our training data and the EviLens benchmark.}
    \label{fig:data showcase}
\vspace{-10pt}
    
\end{figure}

To answer this question, we introduce EviRover, to our knowledge the first agent trained to search both within and beyond the image to resolve perceptual queries.
EviRover moves beyond a single \textit{glance}: at each step, it decides whether further evidence is needed and which action can provide it, before producing a location, a boundary, or a count.
As existing perception datasets do not cover this setting, we design two data generation pipelines, each targeting one form of evidence insufficiency.
For evidence present in the image but not discernible at a glance, we collect high-resolution images with targets occupying less than 0.1\% of the image area, together with scenes requiring multi-step visual exploration, such as I-spy and spot-the-difference puzzles.
For evidence beyond the image, we construct queries over group photographs of public figures and anime characters through two procedures:
(1) starting from group photographs and verifying the identities of the depicted individuals; and (2) starting from a reference image of a known individual or character and synthesizing a group photograph containing them with GPT-Image-2~\citep{openai2026gptimage2}, followed by filtering with Seed-2.0-Pro~\citep{seed2026seed2} for visual quality and identity preservation.
These pipelines yield two training sets, EviRover-SFT-5K and EviRover-RL-12K, and EviLens, a human-verified benchmark of 688 instances spanning segmentation, counting, and three grounding categories: localization, recognition, and spot-the-difference.
Figure~\ref{fig:data showcase} presents representative examples.

Extensive experiments demonstrate the effectiveness of EviRover.
On EviLens, EviRover improves over its Qwen3-VL-4B-Instruct backbone by 30 points on average across the five categories, bringing a 4B model to a level comparable with advanced proprietary models such as Seed-2.1-turbo and Gemini-3.5-Flash.
We also evaluate on WebEyes~\citep{yang2026web}, a benchmark for search-based grounding, segmentation, and visual question answering, where EviRover improves over its backbone by 14.4 IoU on grounding and 24.7 gIoU on segmentation.
The gains further extend beyond our setting: EviRover consistently improves on conventional perception benchmarks such as ReasonSeg~\citep{lai2024lisa} and RefCOCOg~\citep{Mao_2016_CVPR,nagaraja2016modeling}, as well as general multimodal benchmarks, including MMMU~\citep{yue2024mmmu}, MMMU-Pro~\citep{yue2025mmmu}, and MathVerse~\citep{zhang2024mathverse}, with a 15-point gain on BrowseComp-VL~\citep{geng2026webwatcher}.
These results indicate that learning to search under insufficient evidence not only enhances perception but also yields capabilities that generalize well beyond the training distribution.

Our contributions can be summarized as follows:

\begin{itemize}

\item We identify \emph{perception under insufficient evidence} and reformulate perception as an evidence-seeking process that looks beyond a single \textit{glance}.

\item We present \textbf{EviRover}, to our knowledge the first perception agent trained to determine what evidence is missing and how to acquire it.

\item To address the absence of data for perception under insufficient evidence, we design two dedicated data generation pipelines, yielding \textbf{EviRover-SFT-5K}, \textbf{EviRover-RL-12K}, and \textbf{EviLens}, a human-verified benchmark of 688 instances.

\item Extensive experiments validate EviRover: it improves over its backbone by 30 points on average on EviLens and generalizes to WebEyes, conventional perception benchmarks, and general multimodal benchmarks.
\end{itemize}

\section{Related Work}

\paragraph{Visual Perception.}
Grounding~\citep{kamath2021mdetr}, segmentation~\citep{hu2016eccv-segmentation}, and counting~\citep{amini2023open} are conventionally formulated as predicting a location, a boundary, or a count from an image and a query.
Beyond specialized detectors~\citep{liu2024grounding}, segmentation models~\citep{carion2025sam3segmentconcepts}, and counting models~\citep{amini2024countgd}, MLLM-based methods extend these tasks to free-form queries requiring reasoning~\citep{lai2024lisa,you2025seg,liu2025seg}.
In parallel, a growing body of work has focused specifically on fine-grained perception of small objects in high-resolution images~\citep{goto2025referring,zhang2026finers,jia2026small}.
However, these advances retain the one-shot formulation, in which the evidence needed to resolve the query is assumed to be present in the input image.
Our work lifts this assumption, allowing the perceptual prediction to draw on evidence acquired beyond a single glance at the image.

\paragraph{Evidence-Seeking Multimodal Agents.}
A broad class of visual questions cannot be resolved from the image and the model's parametric knowledge alone~\citep{chen-etal-2023-pre-trained,zeng2026vision,choi-etal-2026-progressive}.
One line of work, often referred to as thinking with images~\citep{su2025thinking}, enables models to actively inspect image regions through operations such as zooming and cropping~\citep{zheng2026deepeyes,wu2024v,guo2026thinking,zhang2025thyme}.
Another line develops multimodal search agents that acquire external knowledge through web search, progressing from short-horizon search to long-horizon deep research~\citep{wu-etal-2026-mmsearch,geng2026webwatcher,huang2026vision,yao2026mm}.
However, these works acquire evidence to produce textual answers, treating perception as a means rather than an end.
Recent attempts extend search to segmentation and grounding~\citep{yang2026web,tang2026rose,liang2026segresearch}, but either rely on manually designed prompting or restrict evidence seeking to external search for a specific task.
In contrast, our work treats perception as the goal of evidence seeking, where the acquired evidence must be resolved into a perceptual prediction on the input image.
Since the kind of evidence a query lacks varies, deciding what to acquire and how becomes an integral part of the task.
\section{Method}

\subsection{Dataset Construction}
\label{sec:data}

\begin{figure*}[t]
    \centering
    \includegraphics[width=\textwidth]{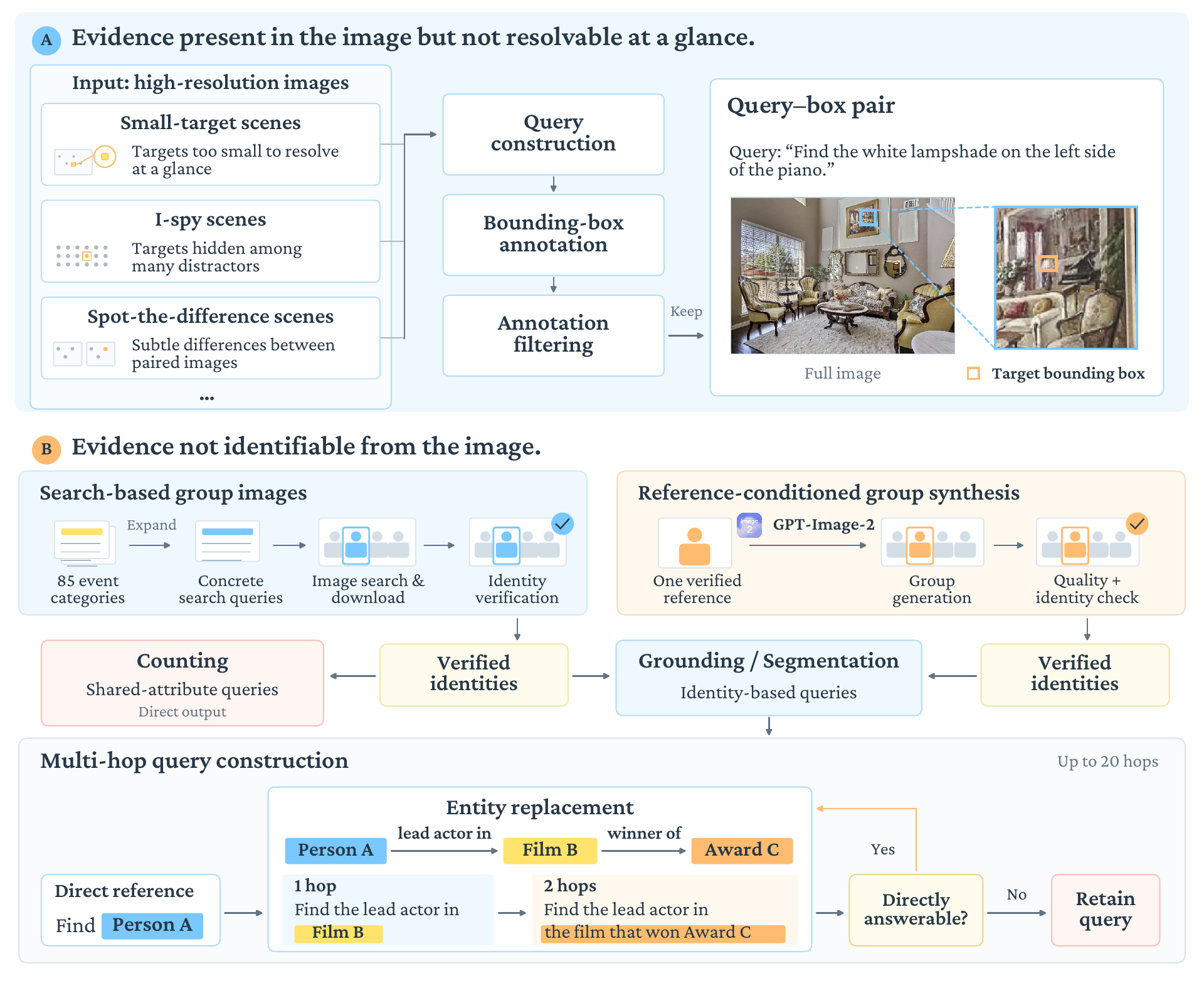}
    \caption{
        Overview of our data generation pipeline.
    }
    \label{fig:data_generation_pipeline}
\end{figure*}

As illustrated in Figure~\ref{fig:data_generation_pipeline}, we design two dedicated data generation pipelines targeting different forms of missing evidence.

\begin{wrapfigure}{r}{0.50\textwidth}
\vspace{-\intextsep}
\centering
\includegraphics[width=\linewidth,trim=0 6 0 0,clip]{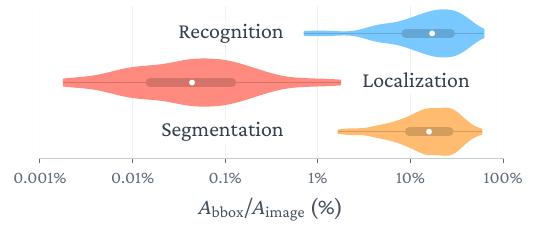}
\setlength{\abovecaptionskip}{2pt}
\caption{Distribution of target area relative to image area.}
\label{fig:target_scale_distribution}
\vspace{-\intextsep}
\end{wrapfigure}

\paragraph{Evidence present in the image but not resolvable at a glance.}
We collect high-resolution images containing targets so small that they cannot be resolved without inspecting the image at a finer scale, 
in many cases occupying well under 0.1\% of the image area, 
together with images where the target can only be found by comparing or scanning multiple regions, 
such as I-spy and spot-the-difference scenes.
Images are adapted from Mini-o3~\citep{lai2026mini} and collected from the web through image search. 

\paragraph{Evidence beyond the image.}
We construct perception queries over group photographs of people, including public figures and anime characters, through two complementary procedures.

The first collects real group photographs and verifies the identities of the people they depict. 
We begin from a manually written seed list of 85 event categories spanning award ceremonies, music groups, esports competitions, political summits, product launches, academic events, sports competitions, and entertainment programs. 
We prompt Seed-1.8 to expand each category into concrete search queries, which are then issued to an image search engine to obtain group photographs and their associated captions.
However, captions are an unreliable source of identity annotations, as they are absent for most collected images and frequently incomplete or inaccurate when present.
We therefore verify identities independently using Seed-2.0-Pro equipped with web search tools.

The second procedure starts from a reference image of a known individual or character and synthesizes a group photograph containing that individual.
Since single-person images with verifiable identities are far more abundant than group photographs in which every identity can be verified, this procedure covers a substantially broader range of identities.
We initially explored conditioning generation on multiple verified reference images simultaneously.
For real individuals, identity preservation degrades sharply as the number of references increases, remaining reliable for only up to approximately three references. 
Beyond this number, the generated individuals can no longer be reliably matched to their references through reverse image search, a failure further confirmed by manual inspection.
Anime characters, in contrast, are defined by distinctive visual traits such as hairstyles and costumes, and remain well preserved under multi-reference conditioning.
We therefore condition each generation of real individuals on a single reference identity while leaving the other individuals in the scene unconstrained, and condition anime generation on multiple character references.
The images are synthesized with GPT-Image-2 and filtered by Seed-2.0-Pro, which assesses visual quality and verifies identity preservation by searching for the generated individual and confirming that the results match the reference identity.

The two procedures thus yield group photographs with different levels of identity coverage: real photographs may have some or all depicted individuals verified, synthesized photographs of real individuals contain exactly one verified individual, and synthesized anime images may contain multiple verified characters.

\paragraph{Query construction.}
The verified identities serve as ground truth for query construction.
Grounding and segmentation queries require at least one verified identity per image, which serves as the target to be localized or segmented.
Counting queries require every depicted individual to be verified, and are formulated over attributes shared among them, such as affiliation with a particular institution or receipt of a particular honor.

In their initial form, these queries name the target directly, for example by asking for the mask of a specific individual.
A model that recognizes the named individual can resolve such queries without seeking any additional evidence.
To make evidence seeking necessary, we rewrite each query into a multi-hop form through iterative entity replacement.
At each step, we select an entity in the current query, obtain its public profile through a search engine, and prompt GPT-5.4-mini to replace the entity with an indirect description that identifies it without naming it.
The replacement terminates once Seed-1.8 can no longer resolve the query without tools, with a maximum of 20 hops.

\begin{wrapfigure}{r}{0.58\textwidth}
\vspace{-\intextsep}
\centering
\includegraphics[width=\linewidth,trim=0 9 0 0,clip]{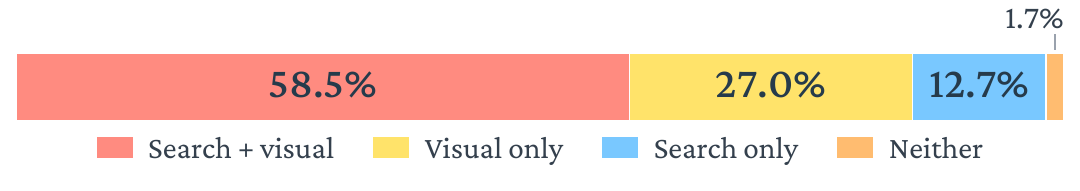}
\setlength{\abovecaptionskip}{0pt}
\caption{Tool-use composition of EviRover-SFT-5K.}
\label{fig:tool_usage}
\vspace{-\intextsep}
\end{wrapfigure}

\paragraph{SFT trajectory construction.}
We generate agent trajectories for the constructed queries, using Gemini-3.0-Pro for localization queries and Seed-2.0-Pro for the remaining tasks, and retain only those that reach the correct answer through a logically consistent sequence of actions.
Correctness alone, however, does not ensure that a trajectory exhibits the intended behavior.
For example, queries about widely recognized individuals are often resolved by a single search or directly from parametric knowledge, involving little evidence seeking.
We therefore downsample such trajectories while retaining a portion of them, so that the model learns both to answer directly when its knowledge suffices and to seek evidence when it does not.
Figure~\ref{fig:tool_usage} shows the resulting composition of EviRover-SFT-5K.

Together, these procedures yield two training sets, EviRover-SFT-5K and EviRover-RL-12K.
% No query is shared between the training sets and EviLens.

% \subsection{EviLens: evaluating perception under insufficient evidence}
% We further build EviLens, a human-verified benchmark of 688 instances for evaluating perception under insufficient evidence, spanning segmentation, counting, and three grounding categories: localization, recognition, and spot-the-difference.
\subsection{EviLens}
\label{sec:evilens}

We construct EviLens with the pipelines above and manually verify every instance, yielding a benchmark of 688 instances for evaluating perception under insufficient evidence.
It covers segmentation, counting, and three grounding categories that differ in the type of missing evidence.
\textit{Localization} targets are explicitly named but tiny or hidden among clutter, as in high-resolution scenes and I-spy puzzles.
\textit{Recognition} targets are visually salient but referred to indirectly, requiring external information to identify.
\textit{Spot-the-difference} targets are defined relative to a second panel and require cross-panel comparison.
\textit{Segmentation} and \textit{counting} share the recognition setting but output masks and counts, respectively.

Table~\ref{tab:benchmark_comparison} compares EviLens with existing benchmarks in task coverage and required abilities.
EviLens contains 140 localization, 182 recognition, 15 spot-the-difference, 195 segmentation, and 156 counting instances, with the spot-the-difference instances comprising 79 annotated differences in total.

\begin{table}[t]
\centering
\caption{
Comparison of EviLens with existing perception benchmarks.
}
\label{tab:benchmark_comparison}
\renewcommand{\arraystretch}{1.25}
\setlength{\tabcolsep}{6pt}
\resizebox{\linewidth}{!}{%
\begin{tabular}{l|ccc|ccc}
\toprule
\multirow{2}{*}{\textbf{Benchmark}}
& \multicolumn{3}{c|}{\textbf{Task}}
& \multicolumn{3}{c}{\textbf{Required ability}} \\
\cmidrule(lr){2-4} \cmidrule(lr){5-7}
& Grounding & Segmentation & Counting
& \shortstack{Visual\\acquisition} & \shortstack{Visual\\comparison} & \shortstack{External\\knowledge} \\
\midrule
RefCOCOg~\citep{Mao_2016_CVPR}    & \yes & \yes & \no  & \no  & \no  & \no  \\
ReasonSeg~\citep{lai2024lisa}      & \no  & \yes & \no  & \no  & \no  & \yes  \\
Ref-Adv~\citep{dong2026ref}        & \yes & \no  & \no  & \no  & \yes & \no  \\
SOREC~\citep{goto2025referring}    & \yes & \no  & \no  & \yes & \no  & \no  \\
OK-VOS~\citep{liang2026segresearch} & \no & \yes & \no  & \no  & \no  & \yes \\
WebEyes~\citep{yang2026web}        & \yes & \yes & \no  & \no  & \no  & \yes \\
\midrule
\textbf{EviLens (Ours)}           & \yes & \yes & \yes & \yes & \yes & \yes \\
\bottomrule
\end{tabular}
}
\end{table}

\paragraph{Metrics.}
Recognition and localization are evaluated by mean box IoU and R@0.5.
Segmentation is evaluated by gIoU (mean per-instance IoU) and cIoU (cumulative intersection over cumulative union).
Counting is evaluated by exact-match accuracy.
Spot-the-difference is evaluated by micro- and macro-F1 under greedy one-to-one matching at IoU 0.5, where micro-F1 pools matches across images and macro-F1 averages per-image scores.

\subsection{Model Training}
\label{sec:training}

We train EviRover in two stages.
SFT establishes the interaction format and basic tool-use skills, while RL teaches the model to identify what evidence each query lacks and how to acquire it, using the correctness of the final perceptual output as feedback.

\paragraph{Agent framework.}
EviRover operates with a set of ten tools, shared across trajectory labeling, training, and evaluation.
Four tools acquire external evidence: text search, text-to-image search, image search, and web browsing.
Three tools support fine-grained visual inspection, such as cropping a region for closer examination and rendering a candidate region on the original image for verification.
The remaining three support task-specific prediction: SAM3-based mask generation, side-by-side region comparison for spot-the-difference queries, and a Python interpreter.
Table~\ref{tab:tool_spec} summarizes the available tools; detailed specifications and complete tool schemas are provided in Appendix~\ref{app:tools}.

\paragraph{Supervised fine-tuning.}
We fine-tune Qwen3-VL-4B-Instruct on EviRover-SFT-5K (Section~\ref{sec:data}) to initialize tool use and task-specific prediction.
The loss is computed only over model-generated tokens, with tool observations masked.

\paragraph{Reinforcement learning.}
We further optimize the SFT model on EviRover-RL-12K.
Since the training data span heterogeneous perception tasks with substantially different reward distributions, we adopt EMA-GRPO~\citep{feng2026onethinker}, which maintains task-specific running reward statistics to normalize the learning signal across tasks.

We use task-specific outcome rewards in $[0,1]$.
Let $\hat p$ denote a predicted box and $g$ its ground-truth counterpart, and let $\hat c$ and $c$ denote the predicted and ground-truth counts.
For segmentation, let $\hat M$ be the mask returned by SAM3 from the predicted box and point prompts, and $M$ the ground-truth mask.
The reward is
\begin{equation}
R=
\begin{cases}
\mathrm{IoU}(\hat p, g), & \text{grounding},\\[4pt]
\mathrm{IoU}(\hat M, M), & \text{segmentation},\\[4pt]
\mathbbm{1}[\hat c = c], & \text{counting},\\[4pt]
F_1^{\mathrm{soft}}(\hat P, G), & \text{spot-the-difference},
\end{cases}
\label{eq:rewards}
\end{equation}
where segmentation predictions are required to provide a bounding box together with at least one positive and one negative point.

For spot-the-difference, a prediction may recover only a subset of the annotated differences, so we use a soft set-level reward.
Let $\hat P=\{\hat p_1,\dots,\hat p_n\}$ be the predicted boxes in the left panel and $G=\{g_1,\dots,g_m\}$ the annotated differences.
We form the candidate set
\begin{equation}
\mathcal C=
\{(i,j):\mathrm{IoU}(\hat p_i,g_j)\ge\tau\},
\end{equation}
sort candidate pairs by IoU in descending order, and greedily construct a one-to-one matching $\mathcal M$.
Each matched pair is weighted by its overlap:
\begin{equation}
\mathrm{STP}
=
\sum_{(i,j)\in\mathcal M}
\mathrm{IoU}(\hat p_i,g_j)^{\alpha},
\qquad
F_1^{\mathrm{soft}}
=
\frac{2\,\mathrm{STP}}{n+m},
\label{eq:soft_f1}
\end{equation}
with the reward taken as zero when $n+m=0$.
The threshold $\tau$ admits partially correct boxes, while $\alpha$ downweights low-quality matches; implementation details are provided in Appendix~\ref{app:rewards}.

For policy-loss aggregation, we use sequence-mean-token-mean reduction instead of the token-mean reduction used in EMA-GRPO. 
This avoids the length bias of token-mean, which linearly amplifies the influence of long rollouts and can cause a small number of unusually long trajectories to disproportionately dominate the policy update.
We discuss this choice in detail in Appendix~\ref{app:token_mean}.

\section{Experiments}

\subsection{Training and Evaluation Settings}
\label{sec:exp.training and eval setting}

\paragraph{Training details.}
We initialize from Qwen3-VL-4B-Instruct and train on a single node of 8 NVIDIA A800-80GB GPUs with FSDP.
Both stages use AdamW ($\beta=(0.9,0.999)$, weight decay $0.01$) with gradient clipping at $1.0$.
Supervised fine-tuning uses a learning rate of $5\times10^{-6}$ under a cosine schedule with warmup ratio $0.03$, a global batch size of $64$, and $2$ epochs.
Reinforcement learning uses a constant learning rate without warmup, and samples $8$ rollouts for each of $8$ prompts per step, giving $64$ rollouts per step divided into $4$ mini-batches for the policy update.

\paragraph{Evaluation details.}
\label{sec:eval_details}
We evaluate primarily on EviLens, and additionally on WebEyes, a search-based grounding, segmentation and VQA benchmark.
All methods receive the same input prompt (Appendix~\ref{app:system prompt}) and use the same task-specific output interfaces, with coordinates normalized to $[0,1000]$ and boxes in \texttt{xyxy} format.
Direct baselines predict a box from the image and query in a single pass; segmentation converts the predicted box into a mask with SAM3, and counting returns an integer.
Agentic baselines and EviRover additionally have access to the tool framework of Section~\ref{sec:training}, with a limit of 35 tool calls per trajectory.
% Because agentic trajectories vary substantially across rollouts, we sample eight rollouts per instance and report the mean.
Further details are given in Appendix~\ref{app:eval details}.

\subsection{Performance Analysis on EviLens}

As shown in Table~\ref{tab:main_results}, EviLens is challenging for all model families.
Among proprietary models, Gemini-3.5-Flash, Seed-2.1-Turbo, and GPT-5.6-Sol perform best overall.
Gemini-3.5-Flash and Seed-2.1-Turbo lead on recognition and counting, and Seed-2.1-Turbo achieves the highest segmentation score of any model ($0.779$ gIoU).
GPT-5.6-Sol performs notably poorly on spot-the-difference, achieving a micro-F1 of only 0.112, as it consistently predicts a single bounding box even when the prompt indicates multiple differences.

\begin{table}[t]
\centering
\caption{Evaluation results on EviLens.}
\label{tab:main_results}
\renewcommand{\arraystretch}{1.3}
\setlength{\tabcolsep}{5pt}
\resizebox{\textwidth}{!}{
\footnotesize
\begin{tabular}{l|cc|cc|cc|cc|c}
\toprule
\multirow{3}{*}{\textbf{Model}}
& \multicolumn{6}{c|}{\textbf{Grounding}}
& \multicolumn{2}{c|}{\textbf{Segmentation}}
& \textbf{Counting} \\
\cmidrule(lr){2-7}\cmidrule(lr){8-9}
& \multicolumn{2}{c|}{Localization}
& \multicolumn{2}{c|}{Recognition}
& \multicolumn{2}{c|}{Spot Diff}
& \multicolumn{2}{c|}{} & \\
\cmidrule(lr){2-3}\cmidrule(lr){4-5}\cmidrule(lr){6-7} %\cmidrule(lr){8-9}
& IoU & R@.5 & IoU & R@.5 & F1\textsubscript{mi} & F1\textsubscript{ma} & gIoU & cIoU & Acc \\
\midrule
\rowcolor{black!10}\multicolumn{10}{l}{\rule[-1pt]{0pt}{9pt}\textit{\textbf{Closed-source Models}}}\\
GPT-5.6-Sol~\citep{openai2026gpt56} & \cellcolor[rgb]{1.000,0.896,0.854}0.358 & \cellcolor[rgb]{1.000,0.876,0.826}0.407 & \cellcolor[rgb]{0.800,0.947,0.853}0.615 & \cellcolor[rgb]{0.794,0.946,0.848}0.660 & \cellcolor[rgb]{0.971,0.958,0.995}0.112 & \cellcolor[rgb]{0.981,0.973,0.997}0.087 & \cellcolor[rgb]{0.803,0.921,0.985}0.627 & \cellcolor[rgb]{0.723,0.889,0.979}0.699 & \cellcolor[rgb]{0.991,0.964,0.765}51.3 \\
GPT-5.6-Luna~\citep{openai2026gpt56} & \cellcolor[rgb]{1.000,0.958,0.942}0.264 & \cellcolor[rgb]{1.000,0.965,0.952}0.250 & \cellcolor[rgb]{0.912,0.977,0.935}0.533 & \cellcolor[rgb]{0.899,0.973,0.926}0.571 & \cellcolor[rgb]{0.918,0.880,0.986}0.192 & \cellcolor[rgb]{0.938,0.909,0.990}0.224 & \cellcolor[rgb]{0.915,0.966,0.994}0.544 & \cellcolor[rgb]{0.915,0.966,0.994}0.564 & \cellcolor[rgb]{0.998,0.993,0.952}35.5 \\
GPT-5.6-Terra~\citep{openai2026gpt56} & \cellcolor[rgb]{1.000,0.939,0.914}0.304 & \cellcolor[rgb]{1.000,0.943,0.920}0.321 & \cellcolor[rgb]{0.923,0.980,0.943}0.528 & \cellcolor[rgb]{0.922,0.980,0.943}0.548 & \cellcolor[rgb]{0.873,0.815,0.979}0.231 & \cellcolor[rgb]{0.760,0.650,0.960}0.313 & \cellcolor[rgb]{0.925,0.970,0.994}0.539 & \cellcolor[rgb]{0.905,0.962,0.993}0.567 & \cellcolor[rgb]{0.996,0.984,0.898}43.4 \\
Seed-2.1-Turbo~\citep{bytedance2026seed21} & \cellcolor[rgb]{1.000,0.927,0.897}0.331 & \cellcolor[rgb]{1.000,0.935,0.909}0.326 & \cellcolor[rgb]{0.718,0.926,0.792}0.636 & \cellcolor[rgb]{0.691,0.919,0.772}0.674 & \cellcolor[rgb]{0.836,0.761,0.973}0.241 & \cellcolor[rgb]{0.890,0.840,0.982}0.273 & \cellcolor[rgb]{0.600,0.840,0.970}0.779 & \cellcolor[rgb]{0.600,0.840,0.970}0.714 & \cellcolor[rgb]{0.984,0.937,0.589}60.5 \\
Seed-1.8~\citep{seed2026seed1} & \cellcolor[rgb]{1.000,0.875,0.825}0.360 & \cellcolor[rgb]{1.000,0.901,0.862}0.393 & \cellcolor[rgb]{0.901,0.974,0.927}0.542 & \cellcolor[rgb]{0.911,0.977,0.934}0.560 & \cellcolor[rgb]{0.931,0.899,0.988}0.190 & \cellcolor[rgb]{0.866,0.805,0.978}0.274 & \cellcolor[rgb]{0.840,0.936,0.988}0.613 & \cellcolor[rgb]{0.847,0.939,0.989}0.610 & \cellcolor[rgb]{0.994,0.977,0.854}48.0 \\
Qwen3.7-Plus~\citep{qwen37plus} & \cellcolor[rgb]{1.000,0.913,0.878}0.351 & \cellcolor[rgb]{1.000,0.846,0.784}0.414 & \cellcolor[rgb]{0.878,0.968,0.910}0.569 & \cellcolor[rgb]{0.875,0.967,0.908}0.606 & \cellcolor[rgb]{0.956,0.936,0.993}0.137 & \cellcolor[rgb]{0.951,0.928,0.992}0.187 & \cellcolor[rgb]{0.901,0.960,0.993}0.577 & \cellcolor[rgb]{0.893,0.957,0.992}0.576 & \cellcolor[rgb]{0.995,0.981,0.875}46.7 \\
Gemini-3.5-Flash~\citep{google2026gemini35flash} & \cellcolor[rgb]{1.000,0.808,0.732}0.373 & \cellcolor[rgb]{1.000,0.808,0.731}0.421 & \cellcolor[rgb]{0.718,0.926,0.792}0.636 & \cellcolor[rgb]{0.746,0.933,0.813}0.668 & \cellcolor[rgb]{0.760,0.650,0.960}0.274 & \cellcolor[rgb]{0.832,0.755,0.972}0.280 & \cellcolor[rgb]{0.889,0.956,0.992}0.579 & \cellcolor[rgb]{0.877,0.951,0.991}0.582 & \cellcolor[rgb]{0.980,0.920,0.480}61.8 \\
\midrule
\rowcolor{black!10}\multicolumn{10}{l}{\rule[-1pt]{0pt}{9pt}\textit{\textbf{Open-Source Segmentation Models}}}\\
Seg-R1-7B~\citep{you2025seg} & -- & -- & -- & -- & -- & -- & \cellcolor[rgb]{0.981,0.993,0.999}0.266 & \cellcolor[rgb]{0.969,0.988,0.998}0.382 & -- \\
Seg-Zero-7B~\citep{liu2025seg} & -- & -- & -- & -- & -- & -- & \cellcolor[rgb]{0.963,0.985,0.997}0.343 & \cellcolor[rgb]{0.953,0.981,0.996}0.407 & -- \\
SAM3-Agent~\citep{carion2025sam3segmentconcepts}  & -- & -- & -- & -- & -- & -- &\cellcolor[rgb]{0.990,0.996,0.999}0.213 & \cellcolor[rgb]{0.984,0.993,0.999}0.287 & -- \\
\midrule
\rowcolor{black!10}\multicolumn{10}{l}{\rule[-1pt]{0pt}{9pt}\textit{\textbf{Open-Source Grounding Models}}}\\
Perception-R1~\citep{yu2026perception} & 0.007 & 0.000 & \cellcolor[rgb]{0.998,1.000,0.999}0.106 & \cellcolor[rgb]{0.998,1.000,0.999}0.044 & 0.000 & 0.000 & -- & -- & 17.3 \\
UniVG-R1~\citep{bai2025univg} & \cellcolor[rgb]{1.000,0.991,0.988}0.051 & \cellcolor[rgb]{1.000,0.994,0.991}0.036 & \cellcolor[rgb]{0.977,0.994,0.983}0.283 & \cellcolor[rgb]{0.977,0.994,0.983}0.264 & \cellcolor[rgb]{0.993,0.989,0.999}0.014 & \cellcolor[rgb]{0.993,0.989,0.999}0.016 & -- & -- & -- \\
\midrule
\rowcolor{black!10}\multicolumn{10}{l}{\rule[-1pt]{0pt}{9pt}\textit{\textbf{Open-Source General Models}}}\\
Minimax-M3~\citep{lai2026minimax} & \cellcolor[rgb]{1.000,0.999,0.998}0.018 & 0.000 & \cellcolor[rgb]{0.971,0.992,0.978}0.286 & \cellcolor[rgb]{0.964,0.991,0.974}0.329 & \cellcolor[rgb]{0.996,0.994,0.999}0.010 & \cellcolor[rgb]{0.996,0.993,0.999}0.013 & \cellcolor[rgb]{0.971,0.988,0.998}0.329 & \cellcolor[rgb]{0.937,0.975,0.995}0.469 & \cellcolor[rgb]{0.998,0.990,0.936}37.8 \\
% Qwen3.6-27B & \cellcolor[rgb]{1.000,0.880,0.833}0.2065 & \cellcolor[rgb]{1.000,0.905,0.868}0.1972 & \cellcolor[rgb]{0.956,0.988,0.967}0.3580 & \cellcolor[rgb]{0.943,0.985,0.958}0.3511 & \cellcolor[rgb]{0.760,0.650,0.960}0.2769 & \cellcolor[rgb]{0.877,0.820,0.979}0.2280 & \cellcolor[rgb]{0.988,0.995,0.999}0.2768 & \cellcolor[rgb]{0.971,0.988,0.998}0.3694 & \cellcolor[rgb]{0.999,0.997,0.982}32.50 \\
% Qwen3.5-397B-A17B & 0.275 & 0.282 & 0.554 & 0.585& 0.1224 & 0.0938 & 0.6109 & 0.5927 & 47.5 \\
OneThinker-8B~\citep{feng2026onethinker} & \cellcolor[rgb]{1.000,0.982,0.975}0.110 & \cellcolor[rgb]{1.000,0.986,0.981}0.086 & \cellcolor[rgb]{0.956,0.988,0.967}0.353 & \cellcolor[rgb]{0.954,0.988,0.966}0.357 & \cellcolor[rgb]{0.983,0.975,0.997}0.065 & \cellcolor[rgb]{0.974,0.963,0.996}0.106 & \cellcolor[rgb]{0.949,0.979,0.996}0.409 & \cellcolor[rgb]{0.962,0.985,0.997}0.387 & \cellcolor[rgb]{0.999,0.997,0.978}25.6 \\
InternVL-3.5-8B~\citep{wang2025internvl3_5} & \cellcolor[rgb]{1.000,0.996,0.995}0.020 & \cellcolor[rgb]{1.000,0.998,0.997}0.007 & 0.093 & 0.033 & 0.000 & 0.000 & 0.081 & 0.081 & \cellcolor[rgb]{1.000,0.999,0.991}20.5 \\
\midrule
\rowcolor{black!10}\multicolumn{10}{l}{\rule[-1pt]{0pt}{9pt}\textit{\textbf{Our Models}}}\\
% Qwen3.5-0.9B & -- & -- & -- & -- & -- & -- & -- & -- & -- \\
% \quad + SFT & -- & -- & -- & -- & -- & -- & -- & -- & -- \\
% Qwen3.5-2B & -- & -- & -- & -- & -- & -- & -- & -- & -- \\
% \quad + SFT & -- & -- & -- & -- & -- & -- & -- & -- & -- \\
Qwen3VL-4B-Inst~\citep{Qwen3-VL} & \cellcolor[rgb]{1.000,0.950,0.930}0.276 & \cellcolor[rgb]{1.000,0.959,0.942}0.251 & \cellcolor[rgb]{0.984,0.996,0.989}0.265 & \cellcolor[rgb]{0.985,0.996,0.989}0.231 & \cellcolor[rgb]{0.998,0.997,1.000}0.005 & \cellcolor[rgb]{0.998,0.997,1.000}0.008 & \cellcolor[rgb]{0.994,0.998,1.000}0.203 & \cellcolor[rgb]{0.989,0.996,0.999}0.279 & \cellcolor[rgb]{1.000,1.000,0.997}19.2 \\
\quad + SFT & \cellcolor[rgb]{1.000,0.844,0.781}0.369 & \cellcolor[rgb]{1.000,0.920,0.888}0.371 & \cellcolor[rgb]{0.844,0.959,0.885}0.595 & \cellcolor[rgb]{0.844,0.959,0.885}0.626 &\cellcolor[rgb]{0.966,0.950,0.994}0.114& \cellcolor[rgb]{0.958,0.939,0.993}0.179& \cellcolor[rgb]{0.872,0.949,0.990}0.584 & \cellcolor[rgb]{0.806,0.923,0.985}0.633 & \cellcolor[rgb]{0.997,0.987,0.913}42.8 \\
\textbf{EviRover(Ours)} &\cellcolor[rgb]{1.000,0.700,0.580}0.444 &\cellcolor[rgb]{1.000,0.700,0.580}0.500 & \cellcolor[rgb]{0.620,0.900,0.720}0.648 & \cellcolor[rgb]{0.620,0.900,0.720}0.687 &  \cellcolor[rgb]{0.943,0.917,0.991}0.171 & \cellcolor[rgb]{0.912,0.872,0.985}0.266& \cellcolor[rgb]{0.728,0.891,0.980}0.686 & \cellcolor[rgb]{0.665,0.866,0.975}0.708 & \cellcolor[rgb]{0.993,0.972,0.815}50.0 \\
% Qwen3VL-8B-Inst & -- & -- & -- & -- & -- & -- & -- & -- & -- \\
% \quad + SFT & -- & -- & -- & -- & -- & -- & -- & -- & -- \\
\bottomrule
\end{tabular}
}
\end{table}

% Float beside the longer analysis; finish wrapping before Section 4.3.
\begin{wraptable}{R}{0.52\textwidth}
\centering
\caption{Evaluation results on the WebEyes benchmark.}
\label{tab:webeyes}
\renewcommand{\arraystretch}{1.15}
\setlength{\tabcolsep}{3.2pt}
\resizebox{\linewidth}{!}{
\begin{tabular}{l|cc|cc|c}
\toprule
\multirow{2}{*}{\textbf{Model}}
& \multicolumn{2}{c|}{\textbf{Grounding}}
& \multicolumn{2}{c|}{\textbf{Segmentation}}
& \multirow{2}{*}{\textbf{VQA}} \\
\cmidrule(lr){2-3}
\cmidrule(lr){4-5}
& IoU & R@.5
& gIoU & cIoU
& \\
\midrule
Doubao-Seed-2.0-Pro
& 35.7 & 44.4
& 61.2 & 43.3
& 65.4 \\

Gemini-3.1-Pro
& 30.5 & 35.1
& 54.6 & 38.8
& 63.8 \\
\midrule

Qwen3VL-4B-Inst
& 25.6 & 29.9
& 24.8 & 21.7
& 36.0 \\

Qwen3VL-8B-Inst
& 26.8 & 32.6
& 35.8 & 25.9
& 36.3 \\

Pixel-Searcher-8B
& 34.1 & 41.3
& 39.1 & 32.4
& \textbf{42.2} \\
\midrule

\textbf{EviRover(Ours)}
& \textbf{40.0} & \textbf{48.5}
& \textbf{49.5} & \textbf{38.0}
& 40.3 \\
\bottomrule
\end{tabular}
}
\end{wraptable}

Open-source models perform substantially worse.
Despite strong results on conventional perception benchmarks, segmentation specialists reach at most 0.343 gIoU, and grounding specialists achieve near-zero IoU on localization.
General-purpose open-source MLLMs exhibit similarly limited performance.
The difficulty of EviLens therefore lies not in perceptual precision but in acquiring the evidence each query lacks, a capability that existing perception benchmarks do not evaluate.

Against this backdrop, EviRover achieves the best localization and recognition results of all evaluated models, with a clear margin on localization ($0.444$ IoU versus $0.373$ for Gemini-3.5-Flash), and ranks second on segmentation behind Seed-2.1-Turbo.
On counting and spot-the-difference, where its backbone scores $19.2$ and $0.005$ respectively, EviRover surpasses most proprietary models, coming within $1.3$ points of GPT-5.6-Sol on counting.
Relative to its backbone, EviRover improves the primary metric of every category, by $30.2$ points on average, bringing a 4B model to a level comparable with the strongest proprietary systems.

\WFclear
\subsection{Performance on WebEyes}

Table~\ref{tab:webeyes} reports results on WebEyes, where baseline results are taken from the original paper and bold marks the best result among open-source models.
EviRover achieves the best grounding performance among all evaluated models, including the proprietary systems, with $40.0$ IoU and $48.5$ R@$0.5$.
On segmentation, it outperforms Pixel-Searcher-8B, a prompt-based workflow built on a larger backbone, by $10.4$ gIoU.
On VQA, a task absent from our training data, it performs comparably to Pixel-Searcher-8B.
These results suggest that the evidence-seeking behavior learned by EviRover transfers beyond EviLens.

\subsection{Generalization beyond EviLens}

\paragraph{Conventional perception.}
To examine whether evidence-seeking training affects standard perception, we evaluate EviRover on grounding and segmentation in ReasonSeg and RefCOCOg (Table~\ref{tab:traditional_perception}).
EviRover improves over its backbone on every split and metric, and outperforms task-specific models on most metrics, such as Seg-Zero-7B on segmentation and Perception-R1 on grounding.
The capabilities acquired through evidence-seeking training thus carry over to conventional perception.

\begin{table}[H]
\centering
\caption{Evaluation results on traditional perception benchmarks.}
\label{tab:traditional_perception}
\renewcommand{\arraystretch}{1.15}
\setlength{\tabcolsep}{4pt}
\resizebox{\textwidth}{!}{
\begin{tabular}{l|cc|cc|ccc|ccc|cc|cc}
\toprule
\multirow{3}{*}{\textbf{Model}}
& \multicolumn{4}{c|}{\textbf{ReasonSeg}}
& \multicolumn{6}{c|}{\textbf{RefCOCOg Grounding}}
& \multicolumn{4}{c}{\textbf{RefCOCOg Segmentation}} \\
\cmidrule(lr){2-5}
\cmidrule(lr){6-11}
\cmidrule(lr){12-15}
& \multicolumn{2}{c|}{Val}
& \multicolumn{2}{c|}{Test}
& \multicolumn{3}{c|}{Val}
& \multicolumn{3}{c|}{Test}
& \multicolumn{2}{c|}{Val}
& \multicolumn{2}{c}{Test} \\
\cmidrule(lr){2-3}
\cmidrule(lr){4-5}
\cmidrule(lr){6-8}
\cmidrule(lr){9-11}
\cmidrule(lr){12-13}
\cmidrule(lr){14-15}
& gIoU & cIoU & gIoU & cIoU & @50 & @75 & @95 & @50 & @75 & @95 & gIoU & cIoU & gIoU & cIoU \\
\midrule
Seg-Zero-7B
& {62.6} & {\underline{62.0}}
& {57.5} & {52.0}
& {-} & {-} & {-}
& {-} & {-} & {-}
& {68.3} & {65.3}
& {68.8} & {66.8} \\
Perception-R1
& {-} & {-}
& {-} & {-}
& {85.6} & {\underline{75.6}} & {\textbf{31.7}}
& {\underline{85.3}} & {\underline{75.9}} & {\textbf{32.7}}
& {-} & {-}
& {-} & {-} \\
Qwen3VL-4B-Inst
& {\underline{63.9}} & {58.9}
& {\underline{60.9}} & {\underline{53.9}}
& {\underline{85.7}} & {74.9} & {28.6}
& {\underline{85.3}} & {75.4} & {29.4}
& {\underline{71.6}} & {\underline{68.8}}
& {\underline{72.3}} & {\underline{69.9}} \\
\textbf{EviRover (Ours)}
& {\textbf{68.7}} & {\textbf{62.3}}
& {\textbf{62.6}} & {\textbf{56.5}}
& {\textbf{86.1}} & {\textbf{76.4}} & {\underline{31.5}}
& {\textbf{85.4}} & {\textbf{76.7}} & {\underline{32.5}}
& {\textbf{73.7}} & {\textbf{71.9}}
& {\textbf{73.7}} & {\textbf{72.3}} \\
\bottomrule
\end{tabular}
}
\end{table}

\paragraph{General multimodal benchmarks.}
Beyond perception, we assess whether the learned behavior benefits broader multimodal capabilities, spanning multidisciplinary reasoning (MMMU and MMMU-Pro), mathematical reasoning (MathVerse), and agentic visual search (BrowseComp-VL), as shown in Table~\ref{tab:generalization}.
EviRover improves over its backbone on all four benchmarks.
On the reasoning benchmarks the gains reach up to $4$ points, while on BrowseComp-VL accuracy more than doubles, rising from $9.5$ to $24.8$.
The consistent gains on reasoning show that specializing in evidence-seeking perception also improves the model's general capabilities, 
and the substantially larger gain on BrowseComp-VL shows that the learned behavior transfers to agentic visual question answering, which requires acquiring external evidence.

\begin{table}[H]
\centering
\caption{
Generalization to unseen benchmarks.
}
\label{tab:generalization}
\renewcommand{\arraystretch}{1.15}
\setlength{\tabcolsep}{3.2pt}
\resizebox{0.76\columnwidth}{!}{
\begin{tabular}{l|c|cc|c|ccc}
\toprule
\multirow{2}{*}{\textbf{Model}}
& \textbf{MMMU}
& \multicolumn{2}{c|}{\textbf{MMMU-Pro}}
& \textbf{MathVerse}
& \multicolumn{3}{c}{\textbf{BrowseComp-VL}} \\
\cmidrule(lr){2-2}
\cmidrule(lr){3-4}
\cmidrule(lr){5-5}
\cmidrule(lr){6-8}
& Acc.
& Std. & Vision
& Overall
& L1 & L2 & Overall \\
\midrule
Qwen3-VL-4B-Inst
& 63.1
& 44.5 & 46.9
& 62.4
& 10.6 & 8.5 & 9.5 \\

\textbf{EviRover (Ours)}
& \textbf{66.2}
& \textbf{48.5} & \textbf{48.4}
& \textbf{64.5}
& \textbf{33.7} & \textbf{16.0} & \textbf{24.8} \\
\bottomrule
\end{tabular}
}
\end{table}

\section{Conclusion}

In this work, we study perception under insufficient evidence, where a single glance at the image does not suffice to resolve the query.
We reformulate perception in this setting as an evidence-seeking process.
To support this formulation, we develop two dedicated data generation pipelines and construct EviLens, a human-verified benchmark for evaluating perception under insufficient evidence.
Building on these data, we train EviRover to actively seek the missing evidence required by each query through supervised fine-tuning followed by agentic reinforcement learning.
Experiments show that EviRover substantially improves over its backbone on EviLens and WebEyes, and that these gains extend to conventional perception and general multimodal benchmarks.
These results suggest that perception can move beyond a single glance: learning to identify and acquire missing evidence improves perceptual accuracy and yields evidence-seeking capabilities that generalize to broader multimodal reasoning.

% Add acknowledgments when appropriate.

\bibliography{references}
\bibliographystyle{references}

\appendix
\newpage
\section{Appendix}

\subsection{Dataset Construction}

We use Gemini-3.0-Pro as the annotation model for localization because, at the time of data annotation, it was the only model that could successfully complete the task.

Beyond correctness, we also filter SFT trajectories by their behavior patterns.
Trajectories in which the annotation model answers directly, or verifies its answer with only a single tool call, typically reflect knowledge that the annotation model already possesses.
Since Qwen3VL-4B-Instruct may lack this knowledge, imitating such trajectories would teach it to answer without supporting evidence, encouraging hallucination.
We therefore downsample these trajectories, retaining a portion so that the model still learns to answer directly when its own knowledge suffices.

\begin{figure}[H]
\centering
\begin{minipage}[t]{0.48\textwidth}
\centering
\includegraphics[width=\linewidth]{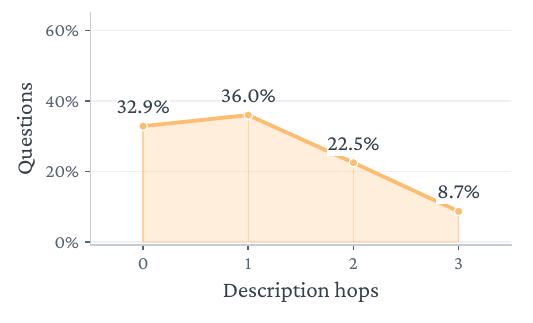}
\small (a) Anime questions
\end{minipage}\hfill
\begin{minipage}[t]{0.48\textwidth}
\centering
\includegraphics[width=\linewidth]{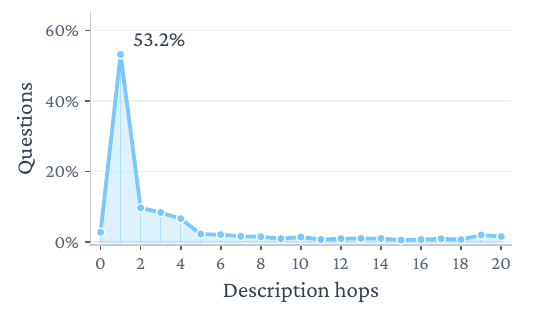}
\small (b) Real-person questions
\end{minipage}
\caption{Distribution of description hops in the RL data for anime and real-person questions.}
\label{fig:rl_hop_distribution}
\end{figure}

Figure~\ref{fig:rl_hop_distribution} shows the distribution of description hops constructed for anime and real-person queries in the RL data.
Anime queries contain substantially fewer hops than real-person queries.
Since entity replacement terminates once Seed-1.8 can no longer resolve the query without tools, this difference reflects the limited coverage of anime characters in the model's parametric knowledge.
For example, a single hop describing a character as released in a specific game at a particular time with a particular skill is often sufficient to terminate the construction.

\subsection{Training Details}
\label{app:training details}

\subsubsection{Reward Details}
\label{app:rewards}

\paragraph{Segmentation fallback.}
The segmentation reward requires SAM3 to return a mask for the predicted box and point prompts.
When SAM3 fails to produce a mask, we do not assign zero reward.
Under group-relative advantage estimation, a zero reward is not a neutral outcome: it enters the group mean and shifts the advantages of every other rollout for the same prompt, so a tool failure unrelated to prediction quality would corrupt the learning signal for the rest of the group.
We instead treat the predicted box as a rectangular mask and compute the IoU against the ground truth, which preserves the ordering of rollouts by localization quality while remaining a strict lower bound on what SAM3 would have returned.
Trajectories scored this way are flagged in the training metadata so that the fallback rate can be monitored.

\paragraph{Spot-the-difference matching.}
We set $\tau=0.1$ and $\alpha=2$.

The candidate threshold $\tau$ is deliberately set below the evaluation threshold of $0.5$. Using the evaluation threshold for reward computation would assign zero credit to predictions that overlap a ground-truth difference but have not yet reached the evaluation criterion, yielding no intermediate reward signal for approximately localized predictions. Setting $\tau=0.1$ instead allows such predictions to receive partial credit and provides a denser learning signal during training.

A low matching threshold alone, however, can make weak overlaps overly rewarding. In particular, if every match above $\tau$ were counted as a unit true positive, a loosely localized box with $\mathrm{IoU}=0.1$ would receive the same positive credit as a tightly localized one with $\mathrm{IoU}=0.9$. This can encourage overprediction, since emitting additional coarse boxes increases the chance of obtaining low-IoU matches. We therefore weight each matched pair by $\mathrm{IoU}^{\alpha}$. With $\alpha=2$, for example, matches at IoU $0.15$ and $0.9$ contribute $0.0225$ and $0.81$, respectively. Thus, weak matches remain informative but contribute substantially less than well-localized predictions.

This weighting also interacts naturally with the set-level precision--recall trade-off. Let the current soft true-positive mass be $S$ and the current reward be

$$
F_1^{\mathrm{soft}}=\frac{2S}{n+m}.
$$

Adding one prediction that matches a previously unmatched ground-truth difference with IoU $q$ changes the reward to

$$
\frac{2(S+q^\alpha)}{n+m+1}.
$$

The additional prediction improves the reward if and only if

$$
q^\alpha > \frac{S}{n+m}
= \frac{F_1^{\mathrm{soft}}}{2}.
$$

Hence, an additional box is beneficial only when the quality of the newly recovered match is sufficiently high relative to the current set-level performance. Unmatched or duplicate predictions increase $n$ without increasing $S$ and therefore strictly decrease the reward. Together, the low threshold $\tau$ and the exponent $\alpha$ provide a dense signal for approximate localization while discouraging indiscriminate or overly coarse box proposals.

Matching is greedy rather than optimal.
Sorting candidate pairs by IoU and accepting each pair whose predicted box and ground-truth region are both unmatched costs $O(nm\log nm)$ and is stable in practice, whereas optimal bipartite matching would change the reward only in rare configurations where a lower-IoU pairing yields a higher total.

\subsection{Evaluation Details}
\label{app:eval details}

\paragraph{Coordinate formats.}
Gemini predicts boxes in \texttt{yxyx} format, as $(y_{\min}, x_{\min}, y_{\max}, x_{\max})$, whereas the remaining models use \texttt{xyxy}, as $(x_{\min}, y_{\min}, x_{\max}, y_{\max})$.
All predictions are converted to a common format before scoring.

\paragraph{Decoding and tools.}
All models are decoded at temperature $0.7$.
Text search, image search, and text-to-image search are served by Serper; web page content is retrieved with Jina.

\paragraph{Spot-the-difference.}
Predicted boxes are first restricted to those lying entirely within the left panel, since the task requires differences to be marked there.
The remaining boxes are matched to the annotated differences by greedy one-to-one assignment: candidate pairs with IoU at least $0.5$ are sorted by IoU in descending order, and each pair is accepted if neither its predicted box nor its annotated region has already been matched.
Each accepted pair counts as a true positive.
Precision, recall, and F1 are then computed from the resulting counts.
We report micro F1, which pools true positives, predictions, and annotated regions across all images before computing F1, and macro F1, which computes F1 per image and averages over images.
Micro F1 is treated as the primary metric: with 15 images containing 79 annotated differences, it is the more stable of the two.
Each instance is evaluated over eight rollouts, and both metrics are averaged across rollouts.

\subsection{Why Token-Mean Reduction Is Ill-Suited for Agentic RL}
\label{app:token_mean}

We consider a group of $N$ trajectories sampled for the same input. 
Let trajectory $i$ contain $L_i$ valid response tokens and receive a trajectory-level advantage $A_i$.
In GRPO, the advantages are centered within each group, such that
\begin{equation}
    \sum_{i=1}^{N} A_i = 0.
    \label{eq:adv_center}
\end{equation}

For clarity, we first omit PPO clipping and importance ratios, since the issue considered here arises purely from the reduction over tokens and trajectories.
Let
\begin{equation}
    g_{i,t}
    =
    \nabla_\theta
    \log \pi_\theta(a_{i,t}\mid s_{i,t})
\end{equation}
denote the score function of token $t$ in trajectory $i$, and define its trajectory-average score as
\begin{equation}
    \bar g_i
    =
    \frac{1}{L_i}
    \sum_{t=1}^{L_i} g_{i,t}.
    \label{eq:traj_avg_grad}
\end{equation}
The same argument applies to the clipped PPO surrogate by replacing $g_{i,t}$ with the corresponding token-level surrogate gradient.

\paragraph{Token-mean induces length-biased trajectory weighting.}
Under token-mean reduction, the policy loss is
\begin{equation}
    \mathcal L_{\mathrm{tok}}
    =
    -
    \frac{1}{\sum_j L_j}
    \sum_{i=1}^{N}
    \sum_{t=1}^{L_i}
    A_i
    \log \pi_\theta(a_{i,t}\mid s_{i,t}).
    \label{eq:token_mean_loss}
\end{equation}
Its gradient can be rewritten as
\begin{equation}
    \nabla_\theta \mathcal L_{\mathrm{tok}}
    =
    -
    \sum_{i=1}^{N}
    \frac{L_i}{\sum_j L_j}
    A_i \bar g_i.
    \label{eq:token_mean_grad}
\end{equation}
Thus, trajectory $i$ receives effective weight
\begin{equation}
    q_i^{\mathrm{tok}}
    =
    \frac{L_i}{\sum_j L_j}.
    \label{eq:token_weight}
\end{equation}

In contrast, sequence-mean-token-mean gives
\begin{equation}
    \mathcal L_{\mathrm{seq}}
    =
    -
    \frac{1}{N}
    \sum_{i=1}^{N}
    \frac{1}{L_i}
    \sum_{t=1}^{L_i}
    A_i
    \log \pi_\theta(a_{i,t}\mid s_{i,t}),
    \label{eq:seq_mean_loss}
\end{equation}
and hence
\begin{equation}
    \nabla_\theta \mathcal L_{\mathrm{seq}}
    =
    -
    \frac{1}{N}
    \sum_i A_i \bar g_i.
    \label{eq:seq_mean_grad}
\end{equation}

Therefore, token-mean does not merely change the normalization constant.
It changes the sampling measure over trajectories from
\begin{equation}
    q_i^{\mathrm{seq}} = \frac{1}{N}
\end{equation}
to the length-biased distribution in Eq.~\eqref{eq:token_weight}.

Equivalently, for any trajectory-level quantity $f_i$,
\begin{equation}
    \mathbb E_{\mathrm{tok}}[f]
    =
    \frac{\sum_i L_i f_i}{\sum_i L_i},
    \label{eq:size_bias_empirical}
\end{equation}
which is the empirical analogue of size-biased sampling,
\begin{equation}
    p_{\mathrm{tok}}(\tau)
    =
    \frac{
        L(\tau)p(\tau)
    }{
        \mathbb E_{p}[L]
    }.
    \label{eq:size_bias_distribution}
\end{equation}

Thus, long trajectories are systematically overrepresented in the policy update.
This distinction is particularly important in agentic RL, where trajectory length is policy-dependent: additional search, repeated inspection, unsuccessful attempts, or redundant interactions all increase $L_i$.
There is generally no reason for such a trajectory to receive a proportionally larger policy update merely because more tokens were generated.

\paragraph{Token-mean breaks the balance induced by centered advantages.}
Equation~\eqref{eq:adv_center} implies that the total positive and negative advantage masses are exactly balanced:
\begin{equation}
    \sum_{A_i>0} A_i
    =
    \sum_{A_i<0} |A_i|.
    \label{eq:adv_balance}
\end{equation}

Sequence-level averaging preserves this balance at the level of trajectory coefficients.
Token-mean instead produces positive and negative coefficient masses
\begin{equation}
    M_+^{\mathrm{tok}}
    =
    \sum_{A_i>0} L_i A_i,
    \qquad
    M_-^{\mathrm{tok}}
    =
    \sum_{A_i<0} L_i |A_i|.
    \label{eq:weighted_mass}
\end{equation}
Their difference is
\begin{equation}
    M_+^{\mathrm{tok}}
    -
    M_-^{\mathrm{tok}}
    =
    \sum_i L_i A_i.
    \label{eq:mass_difference}
\end{equation}

Since $\bar A = 0$,
\begin{align}
    \operatorname{Cov}(L,A)
    &=
    \frac{1}{N}
    \sum_i
    (L_i-\bar L)(A_i-\bar A)
    \nonumber\\
    &=
    \frac{1}{N}
    \sum_i L_i A_i.
\end{align}
Therefore,
\begin{equation}
    \boxed{
    M_+^{\mathrm{tok}}
    -
    M_-^{\mathrm{tok}}
    =
    N\,\operatorname{Cov}(L,A)
    }.
    \label{eq:covariance_mass}
\end{equation}

Consequently,
\begin{equation}
    \operatorname{Cov}(L,A)<0
    \quad\Longrightarrow\quad
    M_-^{\mathrm{tok}}
    >
    M_+^{\mathrm{tok}}.
    \label{eq:negative_mass_dominates}
\end{equation}

Hence, although GRPO explicitly centers trajectory-level advantages, token-mean reintroduces an imbalance whenever trajectory length correlates with advantage.
This effect does not require long trajectories to be negative more frequently.
It is sufficient for negative advantages to have larger magnitude at large $L$, since Eq.~\eqref{eq:covariance_mass} depends on the joint magnitude of $L_iA_i$, rather than only on the sign probability $P(A_i<0\mid L_i)$.

\paragraph{Heavy-tailed trajectory lengths induce gradient concentration.}
Token-mean also reduces the effective number of trajectories contributing to an update.
Using the trajectory weights in Eq.~\eqref{eq:token_weight}, define the effective sample size as
\begin{equation}
    N_{\mathrm{eff}}
    =
    \frac{1}{
        \sum_i
        \left(q_i^{\mathrm{tok}}\right)^2
    }.
\end{equation}
Substituting Eq.~\eqref{eq:token_weight} gives
\begin{equation}
    N_{\mathrm{eff}}
    =
    \frac{
        \left(\sum_i L_i\right)^2
    }{
        \sum_i L_i^2
    }.
    \label{eq:neff_length}
\end{equation}

Let
\begin{equation}
    \bar L
    =
    \frac{1}{N}\sum_i L_i
\end{equation}
and define the squared coefficient of variation
\begin{equation}
    \mathrm{CV}_L^2
    =
    \frac{
        \frac{1}{N}
        \sum_i (L_i-\bar L)^2
    }{
        \bar L^2
    }.
\end{equation}
Since
\begin{equation}
    \frac{1}{N}\sum_i L_i^2
    =
    \bar L^2
    \left(
        1+\mathrm{CV}_L^2
    \right),
\end{equation}
Eq.~\eqref{eq:neff_length} becomes
\begin{equation}
    \boxed{
    N_{\mathrm{eff}}
    =
    \frac{N}{
        1+\mathrm{CV}_L^2
    }
    }.
    \label{eq:neff_cv}
\end{equation}

Thus,
\begin{equation}
    \mathrm{CV}_L^2 \uparrow
    \quad\Longrightarrow\quad
    N_{\mathrm{eff}} \downarrow.
\end{equation}
For heterogeneous or heavy-tailed agentic trajectories, a small number of unusually long rollouts can therefore account for a disproportionately large fraction of the optimization step.

By contrast, sequence-level averaging uses $q_i=1/N$ and yields
\begin{equation}
    N_{\mathrm{eff}}=N.
\end{equation}
Token-mean therefore converts heavy-tailed trajectory length directly into concentrated optimization influence.

\paragraph{Interaction with negative-advantage updates.}
Consider a sampled token $a$ with softmax logits $z_j$ and the unclipped policy-gradient loss
\begin{equation}
    \ell
    =
    -A\log\pi(a).
\end{equation}
Its gradient with respect to the logits is
\begin{equation}
    \frac{\partial \ell}{\partial z_j}
    =
    -A
    \left(
        \mathbbm{1}[j=a]
        -
        \pi(j)
    \right).
    \label{eq:logit_grad}
\end{equation}

For $A<0$, gradient descent decreases the sampled token logit relative to competing tokens.
For any $j\neq a$,
\begin{equation}
    \Delta(z_a-z_j)
    =
    \eta A
    \left(
        1-\pi(a)+\pi(j)
    \right)
    <0.
    \label{eq:relative_logit}
\end{equation}

Hence, negative-advantage updates suppress the sampled action and redistribute probability mass toward alternatives.
When the sampled action already carries relatively high probability, such updates tend to flatten the local policy distribution.

Combined with Eq.~\eqref{eq:negative_mass_dominates}, token-mean disproportionately magnifies such suppressive updates whenever long trajectories carry more negative advantage mass:
\begin{equation}
    L_i\ \text{large},
    \quad
    A_i<0
    \quad\Longrightarrow\quad
    L_i|A_i|\ \text{large}.
\end{equation}

Thus, even a moderate dependence between trajectory length and advantage can be substantially amplified by linear length weighting.

\subsection{Prompt and Tool Specifications}
\label{app:prompt_tools}

We provide the system prompt and tool specifications used by EviRover during
training and evaluation. Unless otherwise stated, the same prompt and tool
interface are used across all perception tasks.

\subsubsection{System Prompt}
\label{app:system prompt}

The following system prompt is used during training and evaluation.

\begin{promptbox}{System prompt}
You are a multimodal agent. Use the provided tools to gather information and analyze images, then answer the question.

# Output Format (strict)

Each assistant turn must be exactly one of:

(1) Reasoning then tool call:
<think>
...
</think>
<tool_call>
{"name": "tool_name", "arguments": {...}}
</tool_call>

(2) Reasoning then final answer:
<think>
...
</think>
<answer>
...
</answer>

(3) Tool call without reasoning (when no thinking is needed):
<tool_call>
{"name": "tool_name", "arguments": {...}}
</tool_call>

Rules:
- One tool call per turn.
- The final answer must be inside <answer>...</answer>.
- Coordinates are normalized to 0-1000.
- Bounding boxes use xyxy format: [x_min, y_min, x_max, y_max].
- Points use [x, y].
- For grounding tasks, the answer is a bounding box: <answer>[x_min, y_min, x_max, y_max]</answer>.
- For segmentation tasks, the answer is a JSON object: <answer>{"boxes":[x_min,y_min,x_max,y_max],"positive_points":[[x,y],[x,y],[x,y]],"negative_points":[[x,y],[x,y],[x,y]]}</answer>.
- For counting tasks, the answer is an integer: <answer>N</answer>.
- For spot-the-difference tasks, the answer is a list of bounding boxes on the LEFT panel: <answer>[[x1,y1,x2,y2], [x1,y1,x2,y2], ...]</answer>.

# Important Notes About the Target

- The target is often SMALL and easy to miss. Pay attention to fine details, small text, subtle attributes, relative position, precise boundaries, and local context.
- Use tools strategically to reduce uncertainty.
- Use `crop` to inspect small or unclear regions.
- Use `verify` to check global placement on the full image: whether the candidate box is on the correct object instance, in the correct relative location, and at the right overall scale.
- Use `verify_part` to check local content and tightness: whether the box really contains the target, includes all of it, and is not too loose.
- Your bounding box must tightly fit the target — avoid overly large boxes. A good bbox should contain the target and little else.
- For spot-the-difference tasks: differences are often SMALL and subtle (colors, shapes, missing/added elements). All bounding boxes must be on the LEFT panel only (x coordinates in range 0-500). Confirm each candidate with `compare_lr` before reporting it, and drop the ones where both panels look the same.
- For counting tasks: use crop to inspect specific persons or regions more closely when faces are small or occluded.

# Tool Use Policy

First, observe the given images carefully. Then, based on your observation and reasoning, use the tools based on what is uncertain:
- Use `crop` when the target is hard to see, small, densely surrounded by other objects, or requires reading local details.
- Use `compare_lr` for spot-the-difference tasks: when you have a candidate difference on the LEFT panel, call `compare_lr` on it to see that region side by side with the same region of the RIGHT panel. If the two sides look the same, that candidate is NOT a difference — drop it and do not include it in your answer.
- Use `verify` when you have a candidate box and want to check whether it is in the correct place on the full image and whether its scale looks reasonable.
- Use `verify_part` when you want to inspect the exact content inside a candidate box and check whether the target is fully included and tightly framed.
- Use `python` when you need to perform precise coordinate calculations, analyze pixel values, or do numerical computation.
- Use `verify_mask` when you need to verify segmentation quality by overlaying a SAM3-generated mask on the image.
- Use `text_search` when you need to search the web for factual information to identify the target.
- Use `text_search_image` when you need to find reference images by text description.
- Use `image_search` when you need to identify objects or people by reverse image search.
- Use `browse` to extract specific details from a webpage when search results are insufficient.

\end{promptbox}

\subsubsection{Tool Specifications}
\label{app:tools}

Table~\ref{tab:tool_spec} summarizes the available tools and their interfaces.
The complete tool schemas are listed below.

\paragraph{Conventions.}
All spatial coordinates are normalized to $[0,1000]$ independently of image resolution.
Boxes are given in \texttt{xyxy} format and points as $[x,y]$.
The query image is referenced as \texttt{ORIGINAL}; images returned by tools are numbered \texttt{IMG\_001}, \texttt{IMG\_002}, \dots\ in the order they appear in the trajectory, and may be passed as inputs to subsequent tool calls.

\begin{table}[H]
\centering
\caption{Tools available to the agent. \emph{Visual} indicates whether the tool returns an image to the model.}
\label{tab:tool_spec}
\renewcommand{\arraystretch}{1.2}
\setlength{\tabcolsep}{4pt}
\small
\begin{tabularx}{\textwidth}{@{}>{\raggedright\arraybackslash}p{0.22\textwidth}|>{\raggedright\arraybackslash}X|>{\raggedright\arraybackslash}p{0.29\textwidth}|c@{}}
\toprule
\textbf{Tool} & \textbf{Function} & \textbf{Parameters} & \textbf{Visual} \\
\midrule
\texttt{text\_search} & Web text search returning titles, snippets, and links & \texttt{query}, \texttt{top\_k} (default 6) & \\\addlinespace[2pt]
\texttt{text\_search\_image} & Image search from a text query & \texttt{query}, \texttt{top\_k} (default 5, range 1--10) & \checkmark \\\addlinespace[2pt]
\texttt{image\_search} & Reverse image search over the full image or a specified region & \texttt{image\_id}, \texttt{boxes} (optional), \texttt{top\_k} (default 3) & \\\addlinespace[2pt]
\texttt{browse} & Opens a URL and extracts content relevant to a query with a summary model & \texttt{url}, \texttt{query} & \\
\midrule
\texttt{crop} & Crops and enlarges a region for closer inspection & \texttt{image\_id}, \texttt{boxes} & \checkmark \\\addlinespace[2pt]
\texttt{verify} & Renders proposed boxes on the original image & \texttt{image\_id}, \texttt{boxes} & \checkmark \\\addlinespace[2pt]
\texttt{verify\_part} & Crops a proposed region for final confirmation & \texttt{image\_id}, \texttt{boxes} & \checkmark \\
\midrule
\texttt{verify\_mask} & Runs SAM3 on a proposed box with optional point prompts and returns the mask overlay & \texttt{image\_id}, \texttt{boxes}, \texttt{positive\_points}, \texttt{negative\_points} & \checkmark \\\addlinespace[2pt]
\texttt{compare\_lr} & Returns corresponding left and right regions side by side & \texttt{image\_id}, \texttt{boxes} & \checkmark \\\addlinespace[2pt]
\texttt{python} & Executes Python for computation, coordinate transformation, and image analysis & \texttt{code} & $\ast$ \\
\bottomrule
\end{tabularx}
\end{table}

\paragraph{Evidence acquisition.}
\texttt{image\_search} accepts an optional box, allowing the agent to isolate a region before searching rather than querying with the full image.
We use Qwen3.5-27B as the summary model in \texttt{browse}.

\paragraph{Visual inspection and confirmation.}
\texttt{crop} and \texttt{verify\_part} both return an enlarged region, but differ in intent: the former is used to examine detail, the latter to confirm a candidate answer.
\texttt{verify} draws proposed boxes on the original image, allowing the agent to assess and refine a localization against its surrounding context.
All three may be called repeatedly within a trajectory.

\paragraph{Task-specific prediction.}
\texttt{verify\_mask} runs SAM3 locally on the proposed box together with any point prompts and returns the resulting mask as an overlay, so that the agent can inspect and revise its prompts before committing to a final answer.
\texttt{compare\_lr} applies only to spot-the-difference queries: given a box in the left panel, it returns the corresponding regions of both panels side by side, separated by a colored divider. Boxes must lie within the left panel.
The \texttt{python} tool is provided with the original image and all tool-returned images as preloaded variables, along with \texttt{numpy}, \texttt{PIL}, \texttt{cv2}, \texttt{scipy}, \texttt{math}, \texttt{json}, and \texttt{copy}. It returns an image only when the executed code produces one.

\paragraph{Tool schemas.}
The following schemas form the tool-specification block of the system prompt.
Line wrapping and indentation are adjusted for readability.

\begin{promptbox}{Tool schemas}
<tools>
{
  "type": "function",
  "function": {
    "name": "text_search",
    "description": "Search the web for one text query. Use one query per call.",
    "parameters": {
      "type": "object",
      "properties": {
        "query": {"type": "string", "description": "One concise search query."},
        "top_k": {"type": "integer", "description": "Number of results, default 10."}
      },
      "required": ["query"]
    }
  }
}

{
  "type": "function",
  "function": {
    "name": "text_search_image",
    "description": "Search web images by text query and return image references with titles, URLs, page URLs, and local paths.",
    "parameters": {
      "type": "object",
      "properties": {
        "query": {"type": "string", "description": "One concise image search query."},
        "top_k": {"type": "integer", "description": "Number of image results, default 5."}
      },
      "required": ["query"]
    }
  }
}

{
  "type": "function",
  "function": {
    "name": "image_search",
    "description": "Reverse image search: upload the given image (optionally cropped to a bbox) and return visually similar images with evidence summaries.",
    "parameters": {
      "type": "object",
      "properties": {
        "image_id": {"type": "string", "description": "Image id to search, usually ORIGINAL."},
        "image_path": {"type": "string", "description": "Local image file path (optional if image_id is given)."},
        "boxes": {"type": "array", "items": {"type": "number"}, "minItems": 4, "maxItems": 4, "description": "Optional normalized 0-1000 bbox to crop before searching."},
        "top_k": {"type": "integer", "description": "Number of results, default 3."}
      },
      "required": ["image_id"]
    }
  }
}

{
  "type": "function",
  "function": {
    "name": "browse",
    "description": "Browse a webpage URL and extract information relevant to the query.",
    "parameters": {
      "type": "object",
      "properties": {
        "url": {"type": "string", "description": "The webpage URL."},
        "query": {"type": "string", "description": "The information to extract from the page."}
      },
      "required": ["url", "query"]
    }
  }
}

{
  "type": "function",
  "function": {
    "name": "crop",
    "description": "Crop a region from an image so you can inspect details. Returns the cropped and enlarged image.",
    "parameters": {
      "type": "object",
      "properties": {
        "image_id": {"type": "string", "description": "Image id, usually ORIGINAL."},
        "boxes": {"type": "array", "items": {"type": "number"}, "minItems": 4, "maxItems": 4, "description": "Bbox [x_min, y_min, x_max, y_max] in normalized coordinates (0-1000)."}
      },
      "required": ["image_id", "boxes"]
    }
  }
}

{
  "type": "function",
  "function": {
    "name": "verify",
    "description": "Draw an unlabeled rectangle on the image for bbox verification. You can call this multiple times to iteratively refine your bbox.",
    "parameters": {
      "type": "object",
      "properties": {
        "image_id": {"type": "string", "description": "Image id, usually ORIGINAL."},
        "boxes": {"type": "array", "items": {"type": "number"}, "minItems": 4, "maxItems": 4, "description": "Bbox [x_min, y_min, x_max, y_max] in normalized coordinates (0-1000)."}
      },
      "required": ["image_id", "boxes"]
    }
  }
}

{
  "type": "function",
  "function": {
    "name": "verify_part",
    "description": "Crop the bbox region for final verification. You can call this multiple times to iteratively refine your bbox.",
    "parameters": {
      "type": "object",
      "properties": {
        "image_id": {"type": "string", "description": "Image id, usually ORIGINAL."},
        "boxes": {"type": "array", "items": {"type": "number"}, "minItems": 4, "maxItems": 4, "description": "Bbox [x_min, y_min, x_max, y_max] in normalized coordinates (0-1000)."}
      },
      "required": ["image_id", "boxes"]
    }
  }
}

{
  "type": "function",
  "function": {
    "name": "verify_mask",
    "description": "Run local SAM3 from the proposed bbox plus optional positive/negative points and return a visual overlay of the generated mask.",
    "parameters": {
      "type": "object",
      "properties": {
        "image_id": {"type": "string", "description": "Image id, usually ORIGINAL."},
        "boxes": {"type": "array", "items": {"type": "number"}, "minItems": 4, "maxItems": 4, "description": "Normalized 0-1000 bbox."},
        "positive_points": {"type": "array", "items": {"type": "array", "items": {"type": "number"}, "minItems": 2, "maxItems": 2}},
        "negative_points": {"type": "array", "items": {"type": "array", "items": {"type": "number"}, "minItems": 2, "maxItems": 2}}
      },
      "required": ["image_id", "boxes"]
    }
  }
}

{
  "type": "function",
  "function": {
    "name": "compare_lr",
    "description": "For spot-the-difference images only. Given ONE box on the LEFT panel, return the left region and the SAME region of the right panel stitched side by side (separated by a magenta divider) at identical size, so the two can be compared directly. This is the only way to see both versions of a location up close: a location at (x,y) on the left corresponds to (x+500,y) on the right, so a single crop containing both would have to span more than half the image.",
    "parameters": {
      "type": "object",
      "properties": {
        "image_id": {"type": "string", "description": "Image id, usually ORIGINAL."},
        "boxes": {"type": "array", "items": {"type": "number"}, "minItems": 4, "maxItems": 4, "description": "Bbox [x_min, y_min, x_max, y_max] normalized 0-1000 on the LEFT panel; x_max must be <= 500."}
      },
      "required": ["image_id", "boxes"]
    }
  }
}

{
  "type": "function",
  "function": {
    "name": "python",
    "description": "Execute Python code to perform calculations, coordinate transformations, or image analysis. Use print() for text output. Assign a PIL Image to `_result_image` to return a visual result. Code with no output (no print, no _result_image) is not allowed. Pre-injected variables: `original_image` (PIL Image), `IMG_001`..`IMG_00N` (PIL Images from previous tool results), `num_images` (int). Libraries available: numpy (as np), PIL.Image, PIL.ImageDraw, math, json, cv2, scipy, copy.",
    "parameters": {
      "type": "object",
      "properties": {
        "code": {"type": "string", "description": "Python code to execute."}
      },
      "required": ["code"]
    }
  }
}
</tools>
\end{promptbox}

\subsection{Agent Trajectories on Training and Benchmark Examples}

\begin{figure}[H]
    \centering
    \includegraphics[width=\linewidth]{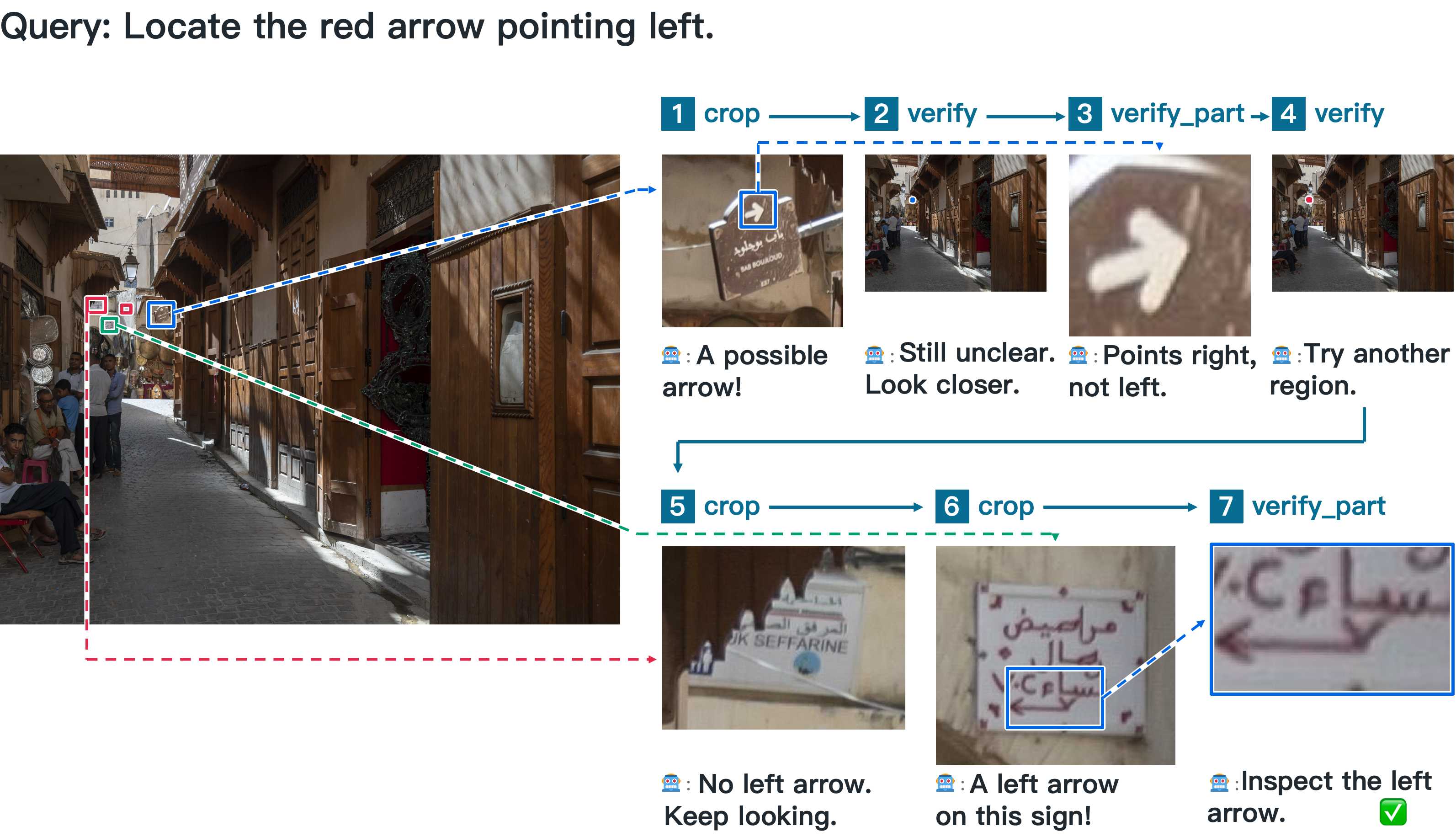}
    \caption{Example localization trajectory generated by EviRover on the training set.}
\end{figure}

\begin{figure}[H]
    \centering
    \includegraphics[width=\linewidth]{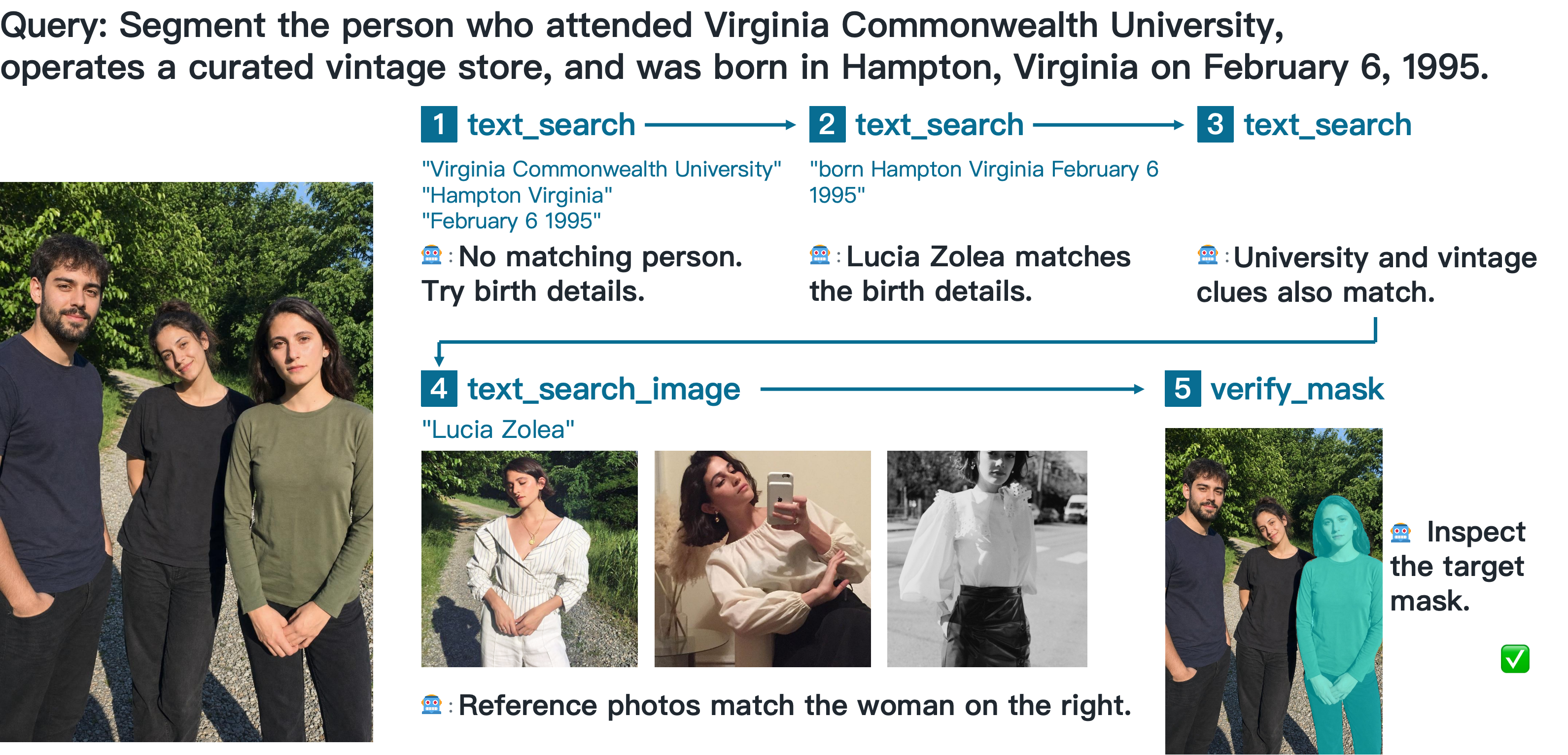}
    \caption{Example segmentation trajectory generated by EviRover on the training set.}
\end{figure}

\end{document}